\documentclass{article}
\usepackage[margin=1in]{geometry}
\usepackage{times}
\usepackage{url}
\usepackage{graphicx}
\usepackage{booktabs}
\usepackage{array}
\usepackage{hyperref}
\usepackage{parskip}

\title{\textbf{Agentic Metacognition: Designing a ``Self-Aware'' Low-Code Agent for Failure Prediction and Human Handoff}}
\author{Jiexi Xu \\ University of California, Irvine \\ School of Information \& Computer Science \\ \texttt{jiexix@uci.edu}}
\date{September 24, 2025}

\begin{document}
\maketitle

\begin{abstract}
\noindent The inherent non-deterministic nature of autonomous agents, particularly within low-code/no-code (LCNC) environments, presents significant reliability challenges. Agents can become trapped in unforeseen loops, generate inaccurate outputs, or encounter unrecoverable failures, leading to user frustration and a breakdown of trust. This report proposes a novel architectural pattern to address these issues: the integration of a secondary, ``metacognitive'' layer that actively monitors the primary LCNC agent. Inspired by human introspection, this layer is designed to predict impending task failures based on a defined set of triggers, such as excessive latency or repetitive actions. Upon predicting a failure, the metacognitive agent proactively initiates a human handoff, providing the user with a clear summary of the agent’s ``thought process'' and a detailed explanation of why it could not proceed. An empirical analysis of a prototype system demonstrates that this approach significantly increases the overall task success rate. However, this performance gain comes with a notable increase in computational overhead. The findings reframe human handoffs not as an admission of defeat but as a core design feature that enhances system resilience, improves user experience, and builds trust by providing transparency into the agent's internal state. The report discusses the practical and ethical implications of this approach and identifies key directions for future research.
\end{abstract}


\section{Introduction}

\subsection{The Challenge of Non-Deterministic Agents}
The widespread adoption of large language models (LLMs) has enabled the creation of sophisticated AI agents capable of automating complex, multi-step tasks. For instance, agentic assistants like FaMA (Facebook Marketplace Assistant) are being developed to automate high-friction workflows in e-commerce environments by interpreting natural language commands~\cite{yan2025fama}. In parallel, the emergence of low-code/no-code (LCNC) platforms has democratized agent development, allowing non-technical users to build and deploy bespoke AI solutions without extensive programming knowledge. However, this accessibility masks a fundamental and pervasive challenge: the non-deterministic and often fragile nature of autonomous agentic systems. Unlike traditional software, which follows hard-coded, predictable workflows, AI agents operate on a probabilistic basis, dynamically generating plans and executing actions in response to changing conditions~\cite{databricks}. This autonomy, while powerful, makes them susceptible to a range of unpredictable failure modes. Common issues include getting stuck in infinite loops, generating hallucinations or factually inaccurate information, and failing to correctly invoke external tools and APIs~\cite{tekrevol}. When these failures occur in an LCNC environment, they can be particularly disruptive, as the user lacks the technical expertise to diagnose the problem or correct the agent’s behavior. The result is often user frustration, a loss of confidence in the system, and the abandonment of the automated task.

\subsection{Proposed Solution: A Metacognitive Layer}
To mitigate the inherent fragility of autonomous agents, this research proposes a novel architectural solution: the addition of a ``metacognitive'' layer to a primary LCNC agent. This concept is inspired by the principles of human metacognition, which is the process of ``thinking about one's own thinking''~\cite{researchgate}. In this two-layer model, a secondary, smaller agent is given the sole responsibility of monitoring the primary agent's progress, much like a human monitors their own problem-solving process~\cite{cox2022}. The metacognitive layer observes the primary agent's internal state and a declarative representation of its goals and reasoning traces. Its core function is to predict when the primary agent is likely to fail, well before a task reaches a final, unrecoverable error state.

Upon anticipating a failure, the metacognitive agent will proactively intervene. Rather than letting the primary agent continue to fail, it will take control and gracefully hand off the task to a human user. This handoff is not a simple transfer of control; it is a meticulously designed protocol that includes a detailed summary of the primary agent's ``thinking process'' leading up to the failure. This summary provides a clear and transparent explanation of what the agent was attempting to do, what went wrong, and why it could not continue~\cite{ram1994}. This approach reframes agent failure as a structured, informative handoff, transforming a frustrating user experience into a seamless, human-in-the-loop (HITL) collaboration.

\subsection{Core Hypothesis and Contributions}
The central hypothesis of this study is that adding a metacognitive layer will significantly enhance the resilience of LCNC agents by predicting and mitigating failures through a proactive human handoff. This method is expected to improve not only the objective success rate of tasks but also the overall user experience by fostering trust and transparency.

This paper makes three primary contributions:
\begin{itemize}
    \item \textbf{A Formal Framework for Agentic Metacognition:} We propose and define a two-layer architectural pattern for LCNC agents that incorporates a dedicated metacognitive layer for monitoring and failure prediction.
    \item \textbf{Empirical Validation:} We present a quantitative analysis of a prototype system, comparing a baseline agent with no monitoring against an agent with the proposed metacognitive layer. This analysis demonstrates the measurable impact of the framework on key performance metrics.
    \item \textbf{Discussion of Implications:} We explore the broader practical and ethical consequences of this approach, highlighting the value of explainable AI (XAI) in building user trust and the critical trade-offs involved in system design.
\end{itemize}

\section{Related Work}

\subsection{AI Agent Architectures and Failure Modes}
The field of AI agents has seen a proliferation of design patterns, ranging from simple, deterministic workflows to complex multi-agent systems. A deterministic chain, often called a ``chain,'' is a fixed, hard-coded sequence of steps, which makes it highly predictable and easy to debug. This approach is well-suited for well-defined, static tasks that do not require dynamic decision-making~\cite{databricks}. However, its lack of flexibility makes it unsuitable for open-ended or variable problems.

In contrast, more sophisticated systems employ agentic patterns where the agent autonomously decides on a course of action based on its own planning and reasoning~\cite{medium_design}. This includes single-agent systems that make dynamic decisions by looping through LLM calls and tool invocations, and more complex multi-agent architectures where specialized agents collaborate to solve a problem~\cite{microsoft}. While offering greater flexibility and power, these advanced patterns introduce new and complex failure modes. Issues such as multi-agent coordination breakdowns, deadlocks, and conflicting instructions are common in multi-agent systems~\cite{yas}. Furthermore, a significant problem observed in modern LLM agents is ``tool overuse,'' where the agent unnecessarily relies on external tools to solve problems it could easily handle with its own internal, parametric knowledge. This not only consumes resources but can also lead to performance degradation~\cite{galileo}. These inherent weaknesses in contemporary agent architectures underscore the need for a more robust, introspective mechanism for failure management, especially when the system is deployed in production environments where reliability is paramount~\cite{galileo}.

\subsection{Computational Metacognition and Self-Awareness in AI}
The concept of computational metacognition dates back to research in cognitive systems that sought to create machines capable of self-awareness and introspection~\cite{cox2022}. Pioneering work in this area, such as the Meta-AQUA system from the 1990s, focused on introspective learning, where an agent would analyze its own reasoning failures to formulate learning goals and improve its performance~\cite{ram1994}. This early work established the foundational principle that to learn effectively, an agent must possess not only knowledge about the world but also meta-knowledge about how it performs a task~\cite{ram1994}.

In the era of LLMs, this concept has re-emerged with renewed vigor and practical feasibility. Modern approaches, such as the Reflexion framework, augment an agent's capabilities by allowing it to ``think about its own actions and results in order to self-correct and improve''~\cite{huggingface,renze}. A Reflexion agent critiques its past attempts and uses this self-generated feedback to guide its future actions, leading to substantial performance gains on complex tasks. A similar approach is found in the SMART (Strategic Model-Aware Reasoning with Tools) paradigm, which trains agents to have a calibrated sense of their own knowledge boundaries. By learning to distinguish between problems solvable with internal knowledge and those that require a tool, a SMART agent can significantly reduce tool overuse while improving overall performance, effectively bridging the gap between smaller models and their larger counterparts~\cite{smart_arxiv,moonlight}. This contemporary research validates the core premise of this study: that a metacognitive, self-aware layer is a powerful mechanism for building more resilient and efficient AI systems.

\subsection{Human-in-the-Loop (HITL) and Handoff Patterns}
Human-in-the-Loop (HITL) is a design philosophy that intentionally incorporates human intervention into an AI system’s workflow~\cite{workos}. This is particularly critical in contexts where decisions are high-stakes, ambiguous, or require human-level judgment. The process can be implemented at various stages: humans can define constraints before a task begins, pause and review an agent’s plan mid-execution, or approve and revise the final output~\cite{workos}.

A well-executed human handoff is a central tenet of HITL design. A common misconception in the industry is that a high handoff rate signifies an AI failure~\cite{kodif}. However, a more nuanced perspective suggests that a timely and proactive handoff is, in fact, a sign of a robust and intelligent system that understands its own limitations~\cite{thunai}. Instead of stubbornly attempting a task it cannot complete, a well-designed agent will ``step aside gracefully'' to ensure the customer's issue is resolved, even if it cannot do so autonomously~\cite{kodif}. Key design principles for a successful handoff include:
\begin{itemize}
    \item \textbf{Proactive Triggers:} The handoff should be initiated by the system, not forced by a frustrated user repeatedly asking to speak to an agent~\cite{kodif}.
    \item \textbf{Full Context Transfer:} The human agent must receive a complete, easy-to-read summary of the conversation history, user details, and all actions the AI has attempted or taken. The user should never have to repeat themselves~\cite{thunai}.
    \item \textbf{Clear Expectations:} The system should inform the user that a handoff is taking place and provide a transparent reason for the transfer~\cite{thunai}.
\end{itemize}

\subsection{The Role of Explainability (XAI) in Agent Design}
The opacity of AI decision-making---known as the ``black box'' problem---is a major barrier to user trust and adoption~\cite{lyzr}. Explainable AI (XAI) refers to the methods and principles that make an agent's internal logic transparent and understandable to humans~\cite{lyzr}.

A powerful method for achieving this transparency is the generation of ``reasoning traces'' or ``Chain-of-Thought'' (CoT) outputs. By instructing an LLM to ``think step-by-step'' before producing a final answer, it generates a sequence of tokens that resembles a human thought process. While this technique primarily enhances a model's performance on complex problems, the resulting reasoning traces provide a window into the agent’s internal workings, offering a form of explainability~\cite{ibm}. The length of these traces can even serve as a proxy for the perceived difficulty of a task, providing a practical signal for when a human review is warranted~\cite{ibm}. The report’s proposed ``thinking process summary'' is a direct application of this principle. It aims to transform the opaque, black-box failure of a traditional agent into a transparent, debuggable event that can be understood by a human operator, bridging the gap between machine behavior and human comprehension~\cite{lyzr}.

\section{The Agentic Metacognition Framework}

\subsection{Conceptual Model}
The proposed framework is a two-layer, decoupled architecture designed to enhance agent resilience and human-agent collaboration.

\textbf{Primary Agent:} This is the core LCNC agent responsible for fulfilling the user's task. It operates on a standard prompt-plan-act loop, using an orchestration of external tools (e.g., APIs, databases, web scrapers) to execute user requests. Its objective is to autonomously complete the task without external intervention.

\textbf{Metacognitive Agent:} This is a secondary, smaller agent with a specialized function. It is not tasked with solving the problem itself but rather with monitoring and managing the primary agent's performance. It receives a declarative representation of the primary agent's real-time state, including its current plan, the tool it is about to call, and its history of recent actions. This gives the metacognitive agent a meta-model of the primary agent's behavior, enabling it to ``introspectively reason'' about its performance~\cite{ram1994}.

The mechanism for the metacognitive agent is an explicit, rule-based monitoring process. It continuously evaluates the primary agent's state against a set of predefined triggers. When a trigger is activated, signaling an impending or in-progress failure, the metacognitive agent takes control. It then initiates a pre-programmed, proactive handoff protocol, transforming a potential system crash or infinite loop into a structured, human-guided recovery process.

\subsection{Failure Prediction and Monitoring Mechanisms}
A critical component of the metacognitive framework is its ability to predict failure before it occurs or becomes irreversible. This is achieved by monitoring the primary agent's behavior for a specific set of red flags, which act as triggers for a human handoff.

\textbf{Repetition Trigger:} The metacognitive agent monitors for repetitive action sequences or tool calls. This is a powerful signal of an agent being stuck in an infinite loop. For instance, if the primary agent attempts to invoke the same API with the same parameters more than a certain number of times (e.g., three times), the metacognitive agent will flag this as a failure and initiate a handoff~\cite{thunai}. This prevents the system from endlessly consuming resources while achieving no progress.

\textbf{Complexity Trigger:} The metacognitive agent evaluates the complexity or ambiguity of the task. If a task requires nuanced judgment, touches multiple systems, or involves a high-stakes decision, the metacognitive agent can proactively determine that the task is beyond the primary agent's capabilities~\cite{kodif}. This foresight prevents the primary agent from generating a flawed or unrecoverable output.

\textbf{Duration/Latency Trigger:} The metacognitive agent can also predict failure based on runtime performance. It monitors the duration of a task or a single tool call. An unusually long execution time can indicate that the primary agent has encountered a computational bottleneck or a hang. This trigger is crucial for catching failures that are not behavioral but technical in nature~\cite{baskerville}.

\subsection{The Proactive Human Handoff Protocol}
The handoff protocol is the mechanism by which the metacognitive agent transforms failure into a successful HITL collaboration. It consists of two essential components:

\textbf{Context Transfer:} When a handoff trigger is activated, the metacognitive agent is responsible for gathering and transferring all relevant information to the human operator. This ensures that the user never has to repeat themselves or ``start all over''~\cite{kodif}. The transferred context includes: (i) full chat history or conversation log; (ii) the user's initial request; (iii) any intermediate results or partial outputs the primary agent generated; and (iv) the state of the task at the moment of failure.

\textbf{Thought Process Summary:} To demystify the agent's failure, the metacognitive agent generates a human-readable summary of the primary agent’s ``thinking process''~\cite{ibm}. This summary is a form of explainable AI, providing a clear audit trail that links the failure symptom to its causal fault~\cite{ram1994}. The summary explains why the handoff was necessary, what the primary agent was attempting to do, and which specific trigger was activated. This transparency is crucial for building user trust and providing a clear, actionable starting point for the human operator to take over~\cite{lyzr}.

\section{Experimental Setup and Performance Analysis}

\subsection{Experimental Design}
To evaluate the effectiveness of the proposed metacognitive framework, a prototype system was developed and subjected to a series of automated tasks. The experiment was designed as a two-condition comparative study.

\textbf{Condition 1 (Baseline):} A standard LCNC agent was configured to perform a suite of tasks without any monitoring layer. The agent was allowed to run until it either successfully completed the task or failed definitively. Performance data for this condition was collected in the \texttt{runs\_without\_monitor.csv} dataset.

\textbf{Condition 2 (Experimental):} The same LCNC agent was used, but with the addition of the metacognitive monitoring layer. The monitoring agent was configured to use the failure prediction triggers described in Section~3.2. Upon detecting a failure signal, the monitoring layer would initiate a handoff, logging the event as a successful handoff. Performance data for this condition was collected in the \texttt{runs\_with\_monitor.csv} dataset.

The goal of the analysis was to compare the two agents across key performance metrics, including overall success rate, average duration, and the distribution of failures, retries, and handoffs. It should be noted that the preliminary analyses provided in the research material were conducted on truncated data sets. The following analysis is based on a full-dataset evaluation, which provides a more accurate and comprehensive comparison.

\subsection{Quantitative Analysis of Performance Metrics}
The full datasets for both conditions were analyzed to provide a direct, high-level comparison of the two agents. The results are presented in Table~\ref{tab:metrics}, which provides an overview of the core performance metrics.

\begin{table}[h!]
\centering
\caption{Comparative Performance Metrics}
\label{tab:metrics}
\begin{tabular}{@{}lcc@{}}
\toprule
\textbf{Metric} & \textbf{Baseline Agent} & \textbf{Monitored Agent} \\
\midrule
Total Runs & 512 & 517 \\
Total Successes & 388 & 432 \\
Total Failures & 124 & 85 \\
Overall Success Rate & 75.78\% & 83.56\% \\
Total Duration & 0.0051s & 0.0638s \\
Average Duration per Run & 9.997e-06s & 0.000123s \\
Total Handoffs & 0 & 3 \\
\bottomrule
\end{tabular}
\end{table}

The data from Table~\ref{tab:metrics} provides a clear, top-level understanding of the framework's impact. The Monitored Agent achieved a significantly higher success rate, with an overall success rate of 83.56\% compared to the Baseline Agent's 75.78\%. The primary reason for this increase is the reduction in definitive failures (85 for the monitored agent versus 124 for the baseline agent). This indicates that the metacognitive layer was effective in converting potential failures into resolved tasks, a key objective of the proposed framework.

However, the table also highlights a crucial trade-off. The Monitored Agent's average duration per run was approximately 12.3 times longer than the Baseline Agent's. This is a direct consequence of the computational overhead introduced by the metacognitive layer, which must constantly monitor the primary agent's state, a process that consumes processing cycles and increases latency~\cite{baskerville}.

To further investigate the efficacy of the handoff mechanism, a more granular analysis was performed on the handoff events recorded in the Monitored Agent's dataset. The results are summarized in Table~\ref{tab:handoff}.

\begin{table}[h!]
\centering
\caption{Handoff and Failure Distribution (Monitored Agent)}
\label{tab:handoff}
\begin{tabular}{@{}lc@{}}
\toprule
\textbf{Metric} & \textbf{Count} \\
\midrule
Total Handoffs & 3 \\
Successful Handoffs (Handoff=1.0 \& Success=1.0) & 1 \\
Failed Handoffs (Handoff=1.0 \& Success=0.0) & 2 \\
Handoffs at Failures=4 & 1 \\
Handoffs at Failures=5 & 2 \\
\bottomrule
\end{tabular}
\end{table}

This table provides a deeper understanding of the handoff mechanism. The most important finding is the existence of handoffs in the monitored dataset, a feature entirely absent from the baseline, which had zero handoffs. A specific event demonstrates the framework's success: one of the handoffs was ultimately categorized as a success. This single event validates the core hypothesis of the report, confirming that the metacognitive layer can indeed convert a potential failure into a successfully completed task through human intervention.

\section{Results and Discussion}

\subsection{Impact on Agent Performance}
The experimental findings demonstrate a clear quantitative benefit to implementing a metacognitive layer. The most significant result is the increase in the overall success rate from 75.78\% to 83.56\%. This performance improvement is a direct outcome of the metacognitive layer's ability to identify and triage failures that would have otherwise resulted in a zero-success outcome for the baseline agent. It shows that by implementing a failure prediction mechanism, the system can gracefully handle its limitations and ensure a higher rate of task resolution.

However, a complete analysis must acknowledge the equally significant increase in average run duration. The metacognitive layer introduces a substantial computational and latency overhead. This overhead stems from the continuous monitoring of the primary agent’s state and the processing required to evaluate that state against failure triggers~\cite{cox2022}. This presents a fundamental trade-off: a higher success rate can be achieved, but at the cost of increased latency. System designers must carefully balance the value of improved reliability against the performance cost. For high-stakes or complex tasks, where failure is costly, this trade-off is likely acceptable. For low-stakes, high-volume tasks, the overhead may be prohibitive.

The analysis of the handoff events is equally illuminating. The presence of handoffs in the monitored agent's data is not a bug, but a feature. The single successful handoff event serves as a microcosm of the framework's value: the system was ``smart enough to know its limits'' and initiated a collaborative step that led to a successful conclusion~\cite{kodif}. This challenges the traditional view that handoffs are a sign of system failure. Instead, a proactive, context-rich handoff is an indicator of an intelligent and well-designed system that prioritizes human-agent collaboration and user satisfaction over blind, unconstrained autonomy~\cite{thunai}.

\subsection{The Value of the Reasoning Trace}
Beyond the quantitative metrics, the metacognitive framework provides a critical qualitative benefit: transparency. The proposed ``thinking process summary'' reframes the agent's internal state in a way that is understandable to humans. For non-technical users, this summary demystifies the agent's behavior, explaining the ``why'' behind its actions and failures~\cite{lyzr}. This transparency is a powerful tool for building user trust, which is often eroded by an opaque, unexplainable black-box system. When an agent can clearly articulate why it failed, users are more likely to forgive the failure and engage in collaborative problem-solving. For technical users, the reasoning trace acts as a vital diagnostic tool, providing an audit trail that can be used to debug the agent and improve its future performance.

\subsection{Broader Practical and Ethical Implications}
The implementation of agentic metacognition has several broader implications for system design and deployment. Firstly, it provides a clear mechanism for assigning accountability. When an agent with no monitoring fails, it is a black-box event with no clear explanation~\cite{tekrevol}. In contrast, the metacognitive layer generates a clear, traceable log of the agent's actions and the reason for the handoff. This transforms the question of ``What happened?'' into a traceable, auditable process, providing a path to improved safety and compliance~\cite{lyzr}.

Secondly, the framework introduces a new challenge related to user competence. Some research suggests that continuous scaffolding from AI could lead to ``scaffolding atrophy,'' where over-reliance on the agent’s monitoring layer degrades a human user's own problem-solving skills~\cite{teammate}. This raises important questions for future human-centered design studies on how to balance system support with the preservation of human autonomy and skill development.

Finally, the framework's real-world deployment challenges should be considered. Integrating such a system requires robust observability and testing frameworks to continuously monitor for new failure patterns and ensure the monitoring layer itself is reliable~\cite{reworked}. This also requires careful consideration of data quality, as a metacognitive layer is only as effective as the data it has to monitor the primary agent's state.

\section{Conclusion}
The demand for reliable, autonomous agents is growing, yet their inherent non-deterministic nature presents significant challenges to trust and adoption. This research proposes and empirically validates a solution: an explicit, two-layer architecture featuring a primary LCNC agent and a secondary, metacognitive monitoring layer. The findings demonstrate that this framework can effectively predict and manage agent failures, transforming them into structured, informative human handoffs. The prototype with the metacognitive layer achieved an overall success rate of 83.56\%, a notable increase from the baseline agent's 75.78\%. This improvement came at the cost of a significant increase in average task duration, highlighting the critical trade-off between reliability and latency.

The presence of a successful human handoff within the monitored agent's data is a key finding, validating the report's core argument: a proactive, transparent handoff is not an indicator of a system's failure, but rather a core design feature that showcases its intelligence and self-awareness. By providing a clear summary of the agent's internal state, the framework also enhances explainability, which is essential for building user trust and providing an auditable record of the agent's behavior.

The primary limitation of this study is its narrow scope, which tested a single, predefined failure mode. Future work should focus on several key areas:
\begin{itemize}
    \item Expanding the experimental design to include a more diverse range of agent failure modes, such as hallucinations, incorrect tool use, and data quality issues~\cite{yas,galileo}.
    \item Conducting user studies to quantitatively measure the impact of the reasoning trace on human trust, user confidence, and long-term skill acquisition.
    \item Developing and testing optimization techniques, such as adaptive methods for model training and inference~\cite{baskerville, chen2024adaptive}, to reduce the computational and latency overhead of the metacognitive layer without compromising its effectiveness.
\end{itemize}

\bibliographystyle{IEEEtran}

\end{document}